\documentclass{article}

\usepackage{microtype}
\usepackage{graphicx}
\usepackage{booktabs} 

\usepackage{hyperref}
\usepackage{fancyhdr}

\usepackage[accepted]{icml2024}

\usepackage{amsmath}
\usepackage{amssymb}
\usepackage{mathtools}
\usepackage{amsthm}

\usepackage[capitalize,noabbrev]{cleveref}

\usepackage{hyperref}
\usepackage{url}
\usepackage{amsmath}

\usepackage{graphicx}
\usepackage{caption}
\usepackage{subcaption}
\usepackage{float}

\usepackage{tabularx}
\usepackage{booktabs}
\usepackage{multirow}
\usepackage{makecell}
\usepackage{tcolorbox}
\usepackage{xcolor}

\newif\ifdraft

\drafttrue

\newcommand{\name}{SWSA}
\newcommand{\pv}[1]{(\textit{p=#1})}

\Crefname{figure}{Fig.}{Figs.}
\Crefname{table}{Tab.}{Tabs.}
\Crefname{section}{Sec.}{Secs.}
\Crefname{proposition}{Prop.}{Props.}
\Crefname{equation}{Eq.}{Eqs.}
\Crefname{appendix}{Supp.}{Supps.}
\Crefname{theorem}{Thm.}{Thms.}
\Crefname{algocf}{Alg.}{Algs.}

\icmltitlerunning{Model Selection of Anomaly Detectors in the Absence of Labeled Validation Data}

\begin{document}

\twocolumn[
\icmltitle{Model Selection of Anomaly Detectors in the Absence of Labeled Validation Data}


\icmlsetsymbol{intern}{*}

\begin{icmlauthorlist}
\icmlauthor{Clement Fung}{cmu,intern}
\icmlauthor{Chen Qiu}{bcai}
\icmlauthor{Aodong Li}{uci}
\icmlauthor{Maja Rudolph}{bcai}
\end{icmlauthorlist}

\icmlaffiliation{cmu}{Software and Societal Systems Department, Carnegie Mellon University, Pittsburgh, PA, USA}
\icmlaffiliation{bcai}{Bosch Center for Artificial Intelligence, Pittsburgh, PA, USA}
\icmlaffiliation{uci}{Department of Computer Science, University of California, Irvine, CA, USA}

\icmlcorrespondingauthor{Clement Fung}{clementf@cs.cmu.edu}

\icmlkeywords{Machine Learning, ICML}

\vskip 0.3in
]



\printAffiliationsAndNotice{\icmlWorkDone} 

\begin{abstract}
    Anomaly detection is the task of identifying abnormal samples in large unlabeled datasets. 
    While the advent of foundation models has produced powerful zero-shot anomaly detection methods, their deployment in practice is often hindered by the absence of labeled validation data---without it, their detection performance cannot be evaluated reliably.
    In this work, we propose SWSA (Selection With Synthetic Anomalies): a general-purpose framework to select image-based anomaly detectors without labeled validation data. Instead of collecting labeled validation data, we generate synthetic anomalies without any training or fine-tuning, using only a small support set of normal images. 
    Our synthetic anomalies are used to create detection tasks that compose a validation framework for model selection. 
    In an empirical study, we evaluate SWSA with three types of synthetic anomalies and on two selection tasks: model selection of image-based anomaly detectors and prompt selection for CLIP-based anomaly detection. 
    SWSA often selects models and prompts that match selections made with a ground-truth validation set, outperforming baseline selection strategies. 
\end{abstract}

\section{Introduction}
Anomaly detection, automatically identifying samples that deviate from normal behavior, is an important technique for supporting medical diagnosis~\cite{fernando2021deep}, safeguarding financial transactions~\cite{Ahmed16}, bolstering cybersecurity~\cite{Mirsky18,Siadati17}, and ensuring smooth industrial operations~\cite{MVTec}.
While there has been significant progress in approaches for unsupervised anomaly detection~\cite{deecke2021transfer,liznerski2022exposing,li2023deep,li2021cutpaste,reiss2021panda}, recent developments in foundation models have made it possible to pre-train a large model on large-scale data from one domain and then to deploy it for a new anomaly detection task. 
In particular, zero-shot CLIP-based anomaly detection approaches~\cite{jeong2023winclip,Li24wacv,Zhou24} have shown great performance in many domains. 
While these approaches provide the exciting possibility to deploy anomaly detectors for new applications without training data, one must trust that they perform as well as expected before deployment. 
Performing such an evaluation is often hindered by a major barrier: the absence of labeled validation data. 
Validation data for anomaly detection is often absent, since anomalies are rare~\cite{gornitz2013toward,trittenbach2021overview}.

\begin{figure*}[t!]
    \centering
    \includegraphics[width=0.825\linewidth]{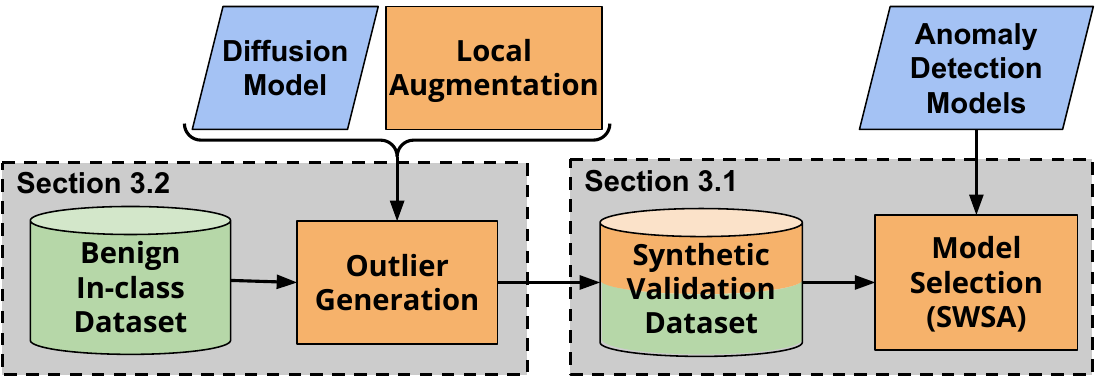}
    \caption{We propose two methods for generating synthetic anomalies, as described in \Cref{sec:anomaly-generation}: image-guided generation with a diffusion model and local augmentation. 
    We produce a synthetic validation set by combining real normal images with synthetic anomalies. 
    The synthetic validation set is then used for model selection, as described in \Cref{sec:anomaly-use}.  
    Components in blue are frozen, components in green are real data, and components in orange are methods implemented in this work.}
    \label{fig:project-diagram}
\end{figure*}

In this work, we study how effective generated synthetic anomalies are for model selection of image-based anomaly detectors through our proposed framework: \name~(Selection With Synthetic Anomalies). 
We compare two promising strategies for generating synthetic anomalies: (i) augmentation methods from prior work~\cite{li2021cutpaste} and (ii) a novel approach that leverages pre-trained diffusion models~\cite{ho2020denoising,song2021denoising,jeong2023training}. 
For a given anomaly detection task, our anomaly generation methods assume access to only a small support set of normal images and 
do not require any training, fine-tuning, or domain-specific techniques. 
We create synthetic validation datasets for \name~by combining synthetic anomalies with normal examples.
When evaluated across a variety of anomaly detection tasks, ranging from natural images to industrial defects, we find that \name~often matches the selections made with real ground-truth validation sets and outperforms baseline selection strategies.
Our work makes the following contributions:
\begin{itemize}
\item In \Cref{sec:anomaly-use}, we propose \name: a framework for selecting anomaly detection models with synthetic anomalies. \Cref{fig:project-diagram} shows the outline of our approach.
\item In \Cref{sec:anomaly-generation}, we propose a practical technique for generating synthetic anomalies with a general-purpose pre-trained diffusion model---without any fine-tuning or auxiliary datasets. When used in \name, we show that these synthetic anomalies are most effective for model selection in natural settings (i.e., birds and flowers).
\item In \Cref{sec:results}, we empirically evaluate \name~with a variety of anomaly-generation methods, datasets, and anomaly-detection tasks. We find that \name~is effective in two use cases: model selection amongst candidate anomaly detectors (\Cref{sec:ad_results}) and prompt selection for zero-shot CLIP-based anomaly detection (\Cref{sec:prompt_selection}).
\end{itemize}

\section{Related Work}

\paragraph{Unsupervised anomaly detection.} 
Recent advances in anomaly detection models include autoencoder-based methods~\cite{chen2018unsupervised,principi2017acoustic}, deep one-class classification~\cite{ruff2018deep,ruff2019deep}, and self-supervised learning-based methods~\cite{bergman2020classification,hendrycks2019using,qiu2021neural,sohn2020learning,ijcai2022p305,schneider2022detecting}. 
While these methods do not require labeled data, their architectures and training frameworks depend on various hyper-parameters, and the choice of hyperparameter can have a strong impact on detection accuracy~\cite{campos2016evaluation,goldstein2016comparative,han2022adbench}. 
Prior work in semi-supervised anomaly detection~\cite{gornitz2013toward,das2016incorporating,trittenbach2021overview,li2023deep} assumes access to a training set with labeled anomalies but obtaining such data is unrealistic for many applications.

In outlier exposure~\cite{hendrycks2018deep}, which has been extended to contaminated data settings~\cite{qiu2022latent} and meta-learning~\cite{li2023zero}, further improvements in detection accuracy come from fine-tuning on an auxiliary dataset, usually Tiny-ImageNet~\cite{tinyimagenet}. 
Although these auxiliary samples provide valuable training signal, we find that they are too dissimilar from normal samples and are easily detected, making them ineffective for model selection. 

\paragraph{Anomaly detection with foundation models.}
Foundation models are pre-trained on massive datasets to learn rich semantic image features and can be effective for anomaly detection in new vision domains with no additional training. Examples include models such as vision transformers~\cite{vit} (ViT) or residual networks~\cite{resnet} (ResNet) pre-trained on the ImageNet dataset~\cite{imagenet}.

Vision-language models, such as CLIP~\cite{radford2021learning}, are another powerful class of foundation models for anomaly detection. With hand-crafted text prompts, CLIP can be employed on a new anomaly detection~\cite{liznerski2022exposing,Zhou24,Li24wacv} or anomaly segmentation task~\cite{jeong2023winclip,zhou2021denseclip} without training data.  
However, detection performance depends on the choice of prompts; a variety of prior work uses prompt learning to find the best-performing prompts for anomaly detection~\cite{Li24cvpr,Li24wacv,jeong2023winclip,Zhou24,esmaeilpour2022zero} but prompt learning also requires additional training or validation data.


\paragraph{Meta-evaluation of anomaly detection.} 
Various prior works~\cite{nguyen2016evaluation,marques2015internal,marques2020internal} propose unsupervised model selection strategies with internal metrics computed from predicted anomaly scores on unlabeled data~\cite{ma2023need}.
Meta-training offers another set of approaches for unsupervised anomaly detector selection~\cite{schubert2023meta,zhao2021automatic,zhao2022toward}. 
However, since meta-learning requires a large number of related labeled datasets, their application has been limited to tabular data, and prior work on internal metrics has been applied to tabular anomaly detection only~\cite{nguyen2016evaluation,marques2015internal,marques2020internal,ma2023need}. 
Shoshan et al.~\cite{shoshan23synthetic} also explore synthetic data for model selection, but their work focuses on optimizing model training; synthetic data is used to select hyper-parameters and stopping criteria. Furthermore, their method trains a GAN to generate synthetic data, which is not applicable to our setting where data is scarce. 

\paragraph{Guided image synthesis.}
Diffusion models~\cite{ho2020denoising,song2021denoising} have recently shown state-of-the-art performance for image synthesis~\cite{dhariwal2021diffusion}. Although these models traditionally generate in-distribution data, prior work guides generative models to generate data from new distributions.
Specifically, text-guided generation~\cite{kim2022diffusionclip,kwon2023diffusion,kawar2023imagic,mokady2023null,tumanyan2023plug} with CLIP~\cite{radford2021learning} guides the image generation process with text prompts to generate samples from a distribution of interest, e.g. ``cat with glasses". 
CLIP embeddings can be used to guide image synthesis in tasks such as style transfer~\cite{karras2019style,karras2020analyzing}, classifier evaluation~\cite{luo2023zero}, and model diagnosis~\cite{jain2023distilling}.
However, these approaches require text labels (e.g, gender, glasses, etc.) as inputs, 
but we assume that the nature of anomalies is unknown.
Instead, we rely on DiffStyle~\cite{jeong2023training}, which performs training-free, guided image generation by interpolating two images during a reverse DDIM process~\cite{song2021denoising}. 
We find that interpolating between two normal samples can preserve dominant visual features (i.e., realistic textures and background) while introducing manipulations that make the generated images promising candidates for \name.

\section{Method}
\label{sec:method}

In this section, we propose to leverage data augmentation and diffusion-based image generation techniques to generate synthetic anomalies. 
By using these synthetic anomalies as a \emph{synthetic validation dataset}, we enable model selection in the absence of a real validation dataset. 
We call our framework \name. In \Cref{sec:anomaly-use}, we describe how synthetic anomalies are used in \name. 
We then propose our synthetic anomaly generation approaches in \Cref{sec:anomaly-generation}. 
\Cref{fig:project-diagram} demonstrates the overall process used in \name.

  


\subsection{Model Selection with Synthetic Anomalies}
\label{sec:anomaly-use}

The absence of labeled validation data is a major roadblock in the deployment of anomaly detection methods; however, normal data can usually be obtained.
For this reason, we follow prior work~\cite{zhao2021automatic} and assume access to a set of normal samples we call the {\em support set} $X_\text{support}$. 
In our empirical study, we show that the support set can have as few as 10 samples. 
We use this support set to construct a synthetic validation set and perform model selection in the following steps:

\noindent {\bf Step 1: Partitioning the support set.} 
    The support set $X_\text{support}$, is randomly partitioned into seed images $X_\text{seed}$ and normal validation images $X_\text{in}$. 
    $X_\text{seed}$ is used for anomaly generation, and $X_\text{in}$ is held out for evaluation.

\noindent {\bf Step 2: Generating synthetic anomalies.} 
    The seed images are processed with either DiffStyle~\cite{jeong2023training} or CutPaste~\cite{li2021cutpaste} to produce synthetic anomalies $\Tilde{X}_\text{out}$. 
    Details are given in \Cref{sec:anomaly-generation}.

\noindent {\bf Step 3: Mixing the synthetic validation set.} 
    $X_\text{in}$ and $\tilde{X}_\text{out}$ are combined to produce a labeled synthetic validation set, 
        \begin{align}
            \mathcal{D} = \{(x, 1) | x \in \Tilde{X}_\text{out} \} \cup  \{(x, 0) | x \in X_\text{in} \}, 
        \end{align}
        where label 1 indicates an anomaly and label 0 indicates a normal image.

\noindent {\bf Step 4: Evaluating detection with candidate models.} 
    We evaluate candidate models by their detection performance on the synthetic validation set $\mathcal{D}$. 
    We use AUROC, the area under the receiving operator characteristic curve, which is typically used to evaluate anomaly detection models~\cite{emmott2015meta}.

       
Since we assume access to only a small support set, training or fine-tuning a generative (diffusion) model is infeasible. 
Instead, in \Cref{sec:anomaly-generation}, we propose to use DiffStyle, a training-free method for diffusion-based image-to-image style transfer, and adapt it to generate synthetic anomalies. 
We also propose to use CutPaste, a data augmentation technique that uses local modifications. 
Our applications of the diffusion-based method and CutPaste do not require any training and do not require additional data beyond $X_\text{seed}$.

\begin{figure*}[tb!]
    \centering
    \begin{subfigure}[b]{0.475\linewidth}
         \centering
         \includegraphics[width=\linewidth]{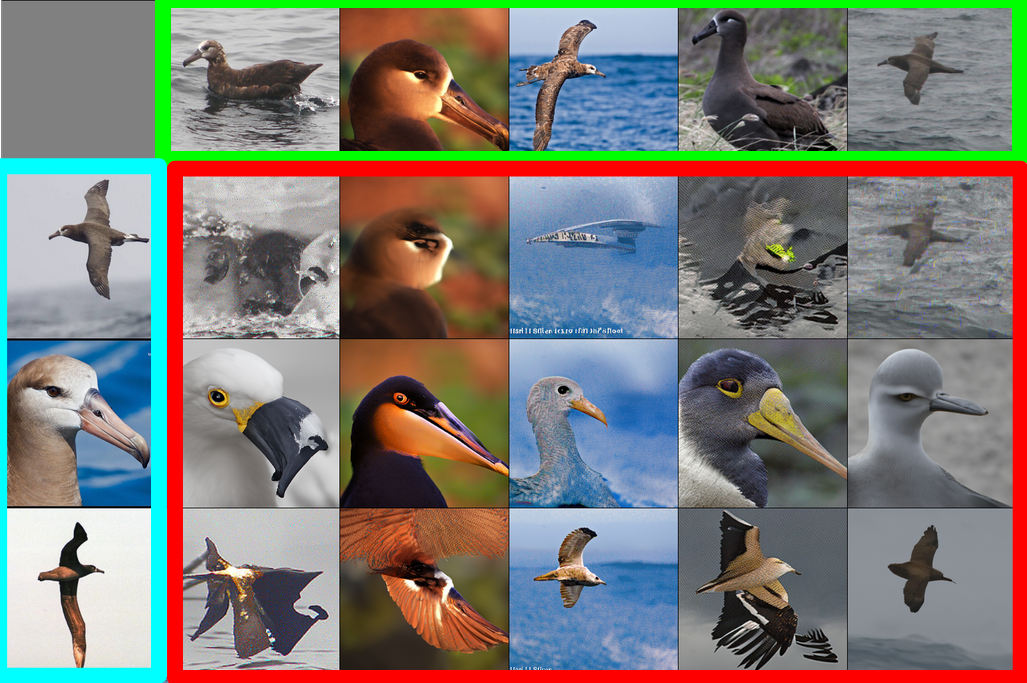}
         \caption{CUB class 1 (``Black Footed Albatross'')}
         \label{fig:cub-grid}
     \end{subfigure}
     \centering
     \begin{subfigure}[b]{0.475\linewidth}
         \centering
         \includegraphics[width=\linewidth]{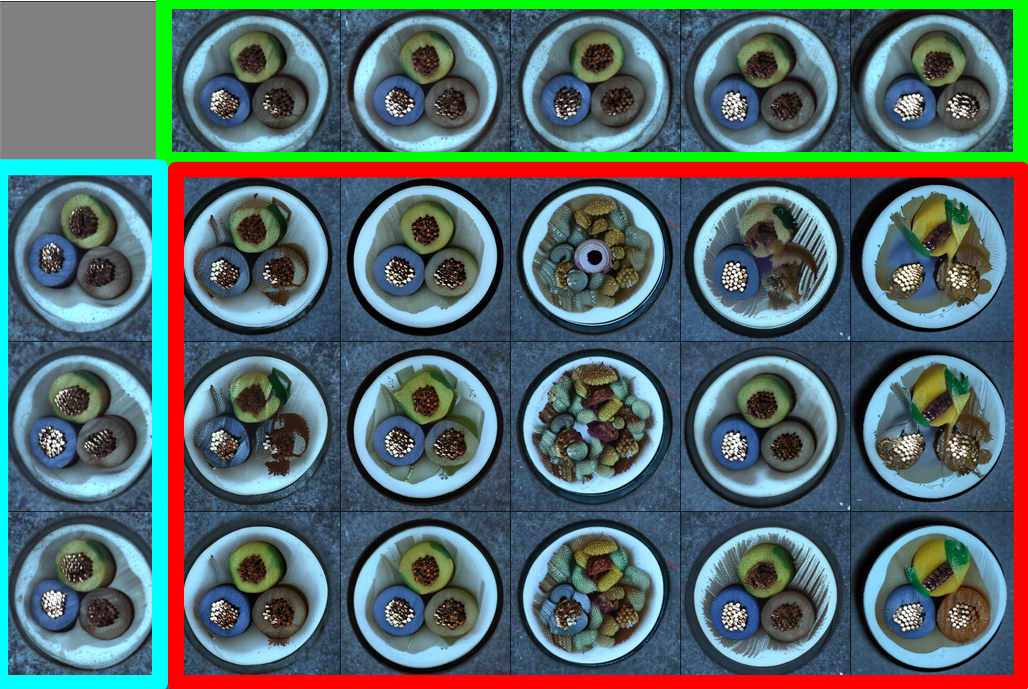}
         \caption{MVTec-AD cable}
         \label{fig:mvtec-grid}
     \end{subfigure}
     \caption{Examples of synthetic anomalies generated with our diffusion-based method for CUB class 1 (left) and MVTec-AD ``cable'' (right). 
     For each example, the top row of images (in green) are used as source ``style'' images, and the left column of images (in cyan) are used as source ``content'' images. 
     The inner grid (in red) shows each pairwise interpolation between the source style and content image, performed with our modified DiffStyle process. 
     All source images are drawn from the distribution of class 1 support images; no validation data or images from other classes are used.}
     \label{fig:interp-examples}
\end{figure*}

\subsection{Generating Synthetic Anomalies}
\label{sec:anomaly-generation}

We propose two methods for generating synthetic anomalies: CutPaste and a diffusion-based method.
Our two methods do not require any training or fine-tuning and only require a small seed set of images $X_\text{seed}$.

\paragraph{CutPaste.} 
CutPaste~\cite{li2021cutpaste} is a data augmentation method that is used for training unsupervised anomaly detection models. 
To modify an image with CutPaste, we randomly select a cropped region of an image and paste it onto a different region of the image.
In our work, we propose a different application for CutPaste's data augmentation: after modifying benign, in-class images from the seed set $X_\text{seed}$, the resulting images are treated as synthetic anomalies for model selection.

\paragraph{Diffusion-based generation.}
We use DiffStyle~\cite{jeong2023training}, diffusion-based style transfer, to generate synthetic anomalies with a pretrained DDIM. 
We first equally divide the seed set $X_\text{seed}$ into style images $X_\text{style}$ and content images $X_\text{content}$. 
DiffStyle takes any style-content image pair $\{I^{(1)}, I^{(2)}\}$ as input---and generates a new image with $I^{(2)}$'s content and $I^{(1)}$'s style. 
To achieve this, $I^{(1)}$ and $I^{(2)}$ are mapped into the diffusion model's latent space through the forward diffusion process to produce latent vectors $x_T^{(1)}$ and $x_T^{(2)}$. 
We refer to the h-space (i.e., the inner-most layer of the UNet) of $x_T^{(1)}$ and $x_T^{(2)}$ as $h^{(1)}$ and $h^{(2)}$ respectively. 
h-space has been shown to be a meaningful semantic space for images, enabling linearity and composition properties, and can be manipulated during a diffusion model's image generation process.
We refer readers for more details related to h-space to their original works~\cite{kwon2023diffusion, jeong2023training}. 

Given two latent vectors $h^{(1)}$ and $h^{(2)}$, we perform a linear interpolation: $h^{(\text{gen})} = (1 - \gamma) h^{(1)} + \gamma h^{(2)}$ where $\gamma$ represents the relative strength of the content image during style transfer.
We use $\gamma = 0.7$ in our experiments.\footnote{The original DiffStyle work implements a spherical interpolation strategy to produce higher-quality images, but we found this was not necessary for our use case.} 
We then perform the asymmetric reverse diffusion process using $x_T^{(1)}$, replacing the h-space with $h^{(\text{gen})}$:
\begin{align}
     x_{t-1} = \sqrt{\alpha_{t-1}} \mathbf{P}_t(\epsilon^{\theta}_t(x_T^{(1)} | h^{(\text{gen})})) + \mathbf{D}_t(\epsilon^{\theta}_t(x_T^{(1)})).
\end{align}

Once the reverse process is completed, the final output $x_0$ is saved as a synthetic anomaly. 
To generate the full set of synthetic anomalies $\tilde{X}_{\text{out}}$, we apply all possibilities of $(I^{(1)}, I^{(2)})$ in the cross product of $X_\text{style}$ and $X_\text{content}$.

We use a pre-trained ImageNet diffusion model~\cite{imagenet, dhariwal2021diffusion} to interpolate between normal images. Our method can be applied even for image domains that are far from ImageNet, such as images of industrial defects~\cite{MVTec,Visa}.
\Cref{fig:interp-examples} shows examples of images generated with this approach; we find that these images have realistic backgrounds and textures, but can contain various semantic corruptions expected of anomalous images.
Our anomaly generation method assumes \emph{no knowledge of the distribution of potential anomalies}: the general-purpose diffusion model is not fine-tuned and only images from the seed set $X_\text{seed}$ are used as inputs.

\section{Empirical Study}
\label{sec:results}


To study the efficacy of synthetic anomalies for model selection, we investigate whether \name~selects similar models and settings as ground-truth validation sets. 
Our evaluation spans various vision domains, including natural and industrial images. 
We first describe the datasets, anomaly detection tasks, and anomaly generation details in \Cref{sec:exp-setup}. 
Next, we showcase two use cases of \name: model selection and CLIP prompt selection---we find that \name~selects the true best-performing model in seven of eight cases (\Cref{sec:ad_results}) and outperforms all other strategies for CLIP prompt selection (\Cref{sec:prompt_selection}); these results are achieved without any access to the real validation data. 

\subsection{Experimental Setup}
\label{sec:exp-setup}

We present an experimental setup that can be used as a benchmark to evaluate synthetic validation data. 
Our benchmark uses a set of anomaly detectors and anomaly detection tasks spanning four datasets. 
The tasks vary by difficulty from the one-vs-rest to the more difficult one-vs-closest setting. 
The goal of this benchmark is to evaluate how well results on synthetic validation data correspond to results one would obtain with ground-truth validation data; we estimate the absolute detection performance, the relative ranking of anomaly detectors, and the optimal hyper-parameters (such as prompts for CLIP-based anomaly detection).

\paragraph{Datasets.}
Our benchmark spans four frequently-used image datasets: Caltech-UCSD Birds (CUB)~\cite{CUB200}, Oxford Flowers~\cite{Flowers102}, MVTec Anomaly Detection (MVTec-AD)~\cite{MVTec}, and Visual Anomaly Detection (VisA)~\cite{Visa}. 
These datasets span both the natural image domain (CUB and Flowers) and the industrial anomaly domain (MVTec-AD and VisA). CUB and Flowers are multi-class datasets containing 200 bird species and 102 flower species respectively. 
MVTec-AD and VisA contain several (i.e., 15 in MVTec-AD, 12 in VisA) real industrial product categories; for each category, the training subset contains images of defect-free products, and the testing subset contains labeled images of both good and defective products. 

\paragraph{Anomaly detection tasks.} 
Our benchmark contains 329 anomaly detection tasks: 15 from MVTec-AD, 12 from VisA, 200 from CUB, and 102 from Flowers. 
For all datasets, each class or product is treated as normal for an individual anomaly detection task. 
In addition to the one-vs-rest anomaly detection setup for multi-class datasets CUB and Flowers, we also adopt the near-anomaly-detection setup used by Mirzaei et al.~\cite{mirzaei2023fake} to simulate more difficult anomaly detection tasks. 
Specifically, after individually selecting each class as the inlier class, we consider each out-class individually and report the class with the worst performance (i.e., one-vs-closest). 
For MVTec-AD and VisA, we predict if an image contains a defective product. 
Each product contains multiple types of defects; we consider all defect types as a single anomalous class when evaluating the one-vs-rest performance, and we consider the worst-performing defect type when reporting the one-vs-closest performance. 
For all tasks, images from the in-class training subset are used as the support set, and images from the relevant in-class and out-class validation subsets are used to construct the ground-truth validation set. 

\paragraph{Generating synthetic anomalies.}
For all four datasets and all 329 anomaly detection tasks, we generate synthetic anomalies with training examples from the in-class distribution only. 
Both the diffusion-based method and CutPaste as used to generate the same number of images with $X_\text{seed}$. 
For the CUB, VisA, and MVTec-AD datasets, we sample 20 images for $X_{\text{seed}}$ from the training set, generating 100 synthetic anomalies with each method. 
For the Flowers dataset, only 10 images are included in the training set for each class, so we generate 25 synthetic anomalies with each method.
\Cref{fig:interp-examples} shows 15 examples of generated synthetic anomalies for a single CUB class (left) and MVTec-AD product (right). 

Although prior results~\cite{kwon2023diffusion,kim2022diffusionclip} suggest that a diffusion model trained in the same domain is required to generate high-quality images, we find that using one common diffusion model can generate sufficiently effective anomalies for \name. 
We use a pre-trained diffusion model (256x256 model trained on ImageNet without class conditioning) from prior work~\cite{dhariwal2021diffusion} for all datasets and anomaly detection tasks.

\subsection{Model Selection with Synthetic Data} 
\label{sec:ad_results}

We first demonstrate the effectiveness of \name~for model selection. Given a set of candidate models, we show that \name~can select the true optimal model.

\paragraph{Candidate anomaly detection models.} 
We experiment across five pre-trained ResNet models (ResNet-152, ResNet-101, ResNet-50, ResNet-34, ResNet-18) and five pre-trained Vision Transformers (ViT-H-14, ViT-L-32, ViT-L-16, ViT-B-32, ViT-B-16). 
For all models, we use the ImageNet pre-trained model weights from prior work~\cite{vit,resnet}. 

\paragraph{Deep-nearest-neighbor anomaly detection.}
To perform anomaly detection, we use the nearest-neighbor-based method of Bergman et al.~\cite{bergman2020deep}. 
We use the values of a candidate model's penultimate layer as the output of a feature extractor $F$ and process the support set $X_\text{support}$ with $F$ to establish a feature bank $Z$:
\begin{align} 
    z_s &= F(x_s), \forall x_s \in X_\text{support}
\end{align}
To perform anomaly detection on an input example $d$, the Euclidean distance between $F(d)$ and its $k$-nearest neighbors in $Z$ is used as an anomaly score $s$:
\begin{align} 
    s &= \sum_{z_s \in Z_k(d)} || F(d) - z_s ||_2^2
\end{align}
where $Z_k(d)$ are the $k$-nearest neighbors to $d$ in the feature bank. 
Bergman et al.~\cite{bergman2020deep} find that small values of $k$ are effective, and we use $k=3$.

\paragraph{Evaluation setup.} 
For each task, we compute the AUROC for each candidate model on both the synthetic validation and real validation datasets. 
We average the AUROC across tasks to compute the ``synthetic AUROC''. 
We repeat this process with real validation datasets, again averaging over tasks to compute the ``real validation AUROC''. 
To evaluate our model selection, we select the model with the best synthetic AUROC for a given task and report the selected model's corresponding real validation AUROC. 
We also compare real and synthetic AUROCs to investigate if the rankings of candidate models are similar.

To establish a baseline for this evaluation, we compare \name~using our synthetic anomalies to \name~using the Tiny-ImageNet dataset. Prior work uses Tiny-Imagenet for outlier exposure~\cite{hendrycks2018deep}, so we investigate if auxiliary examples are effective for model selection. 
When sampling anomalies from Tiny-Imagenet, we sample uniformly at random to generate a dataset $\tilde{X}_\text{out}$ of the same size: 100 images for tasks with CUB, VisA, and MVTec-AD; 25 images for tasks with the Flowers dataset.

\begin{table*}[tb!]
    \centering
    \small
	\caption{When using synthetic anomalies, we report both the accuracy in picking the best model/prompt (``pick rate'') and the resulting AUROC. 
    As an upper bound, we show the AUROC when always picking the best model/prompt (in grey).
    For all settings, \name~picks the best model/prompt more often and produces higher AUROCs than baseline strategies (largest model, default prompt, or prompt ensemble). In particular, \name~is effective on CUB and Flowers with diffusion-based synthetic anomalies.
    }
	\label{table:merged}
	\begin{tabular}{rr |cc cc cc cc}

	\toprule
    &  
    & \multicolumn{2}{c}{\makecell[c]{CUB}} 
    & \multicolumn{2}{c}{\makecell[c]{Flowers}} 
    & \multicolumn{2}{c}{\makecell[c]{MVTec-AD}} 
    & \multicolumn{2}{c}{\makecell[c]{VisA}} \\
    & & Pick rate & AUROC & Pick rate & AUROC & Pick rate & AUROC & Pick rate & AUROC \\
    \midrule
	\multirow{5}{*}{\makecell[c]{One-vs-Closest\\(model selection)}} 
    & Largest model & 11 / 200 & 0.653 & 43 / 102 & 0.956 & 4 / 15 & \textbf{0.716} & 1 / 12 & 0.636 \\
    & SWSA (TinyImg) & 30 / 200 & 0.674 & 50 / 102 & 0.966 & 4 / 15 & \textbf{0.716} & 1 / 12 & 0.636 \\
    & SWSA (Diffusion) & \textbf{66 / 200} & 0.737 & \textbf{59 / 102} & \textbf{0.967} & 4 / 15 & 0.670 & 0 / 12 & 0.643 \\
    & SWSA (CutPaste) & 60 / 200 & \textbf{0.743} & 37 / 102 & 0.945 & 2 / 15 & 0.678 & 1 / 12 & \textbf{0.674} \\
    & Best Model & -- & \textcolor{gray}{0.826} & -- & \textcolor{gray}{0.990} & -- & \textcolor{gray}{0.785} & -- & \textcolor{gray}{0.752} \\
    \midrule
	\multirow{5}{*}{\makecell[c]{One-vs-Rest\\(model selection)}} 
    & Largest model & 32 / 200 & 0.982 & 49 / 102 & \textbf{0.994} & 4 / 15 & \textbf{0.733} & 1 / 12 & 0.765 \\
    & SWSA (TinyImg) & 59 / 200 & 0.982 & 57 / 102 & 0.993 & 4 / 15 & 0.733 & 1 / 12 & 0.765 \\
    & SWSA (Diffusion) & \textbf{109 / 200} & \textbf{0.988} & \textbf{62 / 102} & \textbf{0.994} & 2 / 15 & 0.706 & 0 / 12 & 0.764 \\
    & SWSA (CutPaste) & 62 / 200 & 0.966 & 37 / 102 & 0.974 & 2 / 15 & 0.717 & 1 / 12 & \textbf{0.772} \\
    & Best Model & -- & \textcolor{gray}{0.991} & -- & \textcolor{gray}{0.997} & -- & \textcolor{gray}{0.757} & -- & \textcolor{gray}{0.824} \\
    \midrule
    \multirow{6}{*}{\makecell[c]{One-vs-Closest\\(prompt selection)}} 
      & Default Prompt & 5 / 200 & 0.571 & 1 / 102 & 0.697 & 2 / 15 & 0.741 & 0 / 12 & 0.596 \\
      & Prompt Ensemble & -- & 0.577 & -- & 0.708 & 0 / 15 & 0.728 & 0 / 12 & 0.596 \\
      & SWSA (TinyImg) & 34 / 200 & 0.582 & 18 / 102 & 0.718 & 2 / 15 & 0.760 & 1 / 12 & \textbf{0.612} \\
      & SWSA (Diffusion) & \textbf{46 / 200} & \textbf{0.590} & \textbf{38 / 102} & \textbf{0.729} & 2 / 15 & 0.725 & 2 / 12 & 0.596 \\
      & SWSA (CutPaste) & 34 / 200 & 0.585 & 18 / 102 & 0.718 & 3 / 15 & \textbf{0.763} & 2 / 12 & 0.604 \\
      & Best Prompt & -- & \textcolor{gray}{0.625} & -- & \textcolor{gray}{0.759} & -- & \textcolor{gray}{0.845} & -- & \textcolor{gray}{0.702} \\
    \midrule
    \multirow{6}{*}{\makecell[c]{One-vs-Rest\\(prompt selection)}} 
      & Default Prompt & 0 / 200 & 0.971 & 1 / 102 & 0.959 & 2 / 15 & 0.752 & 0 / 12 & 0.724 \\
      & Prompt Ensemble & -- & 0.972 & -- & 0.962 & 0 / 15 & 0.753 & 0 / 12 & \textbf{0.747} \\
      & SWSA (TinyImg) & 27 / 200 & 0.972 & 13 / 102 & \textbf{0.967} & 5 / 15 & \textbf{0.765} & 1 / 12 & 0.745 \\
      & SWSA (Diffusion) & \textbf{64 / 200} & \textbf{0.973} & \textbf{38 / 102} & \textbf{0.967} & 1 / 15 & 0.728 & 2 / 12 & 0.724 \\
      & SWSA (CutPaste) & 27 / 200 & 0.972 & 22 / 102 & 0.963 & 1 / 15 & 0.746 & 2 / 12 & 0.730 \\
      & Best Prompt & -- & \textcolor{gray}{0.976} & -- & \textcolor{gray}{0.971} & -- & \textcolor{gray}{0.786} & -- & \textcolor{gray}{0.801} \\
    \bottomrule
    
\end{tabular}
\end{table*}

\begin{figure*}[tb!]
    \centering
    \includegraphics[width=0.975\linewidth]{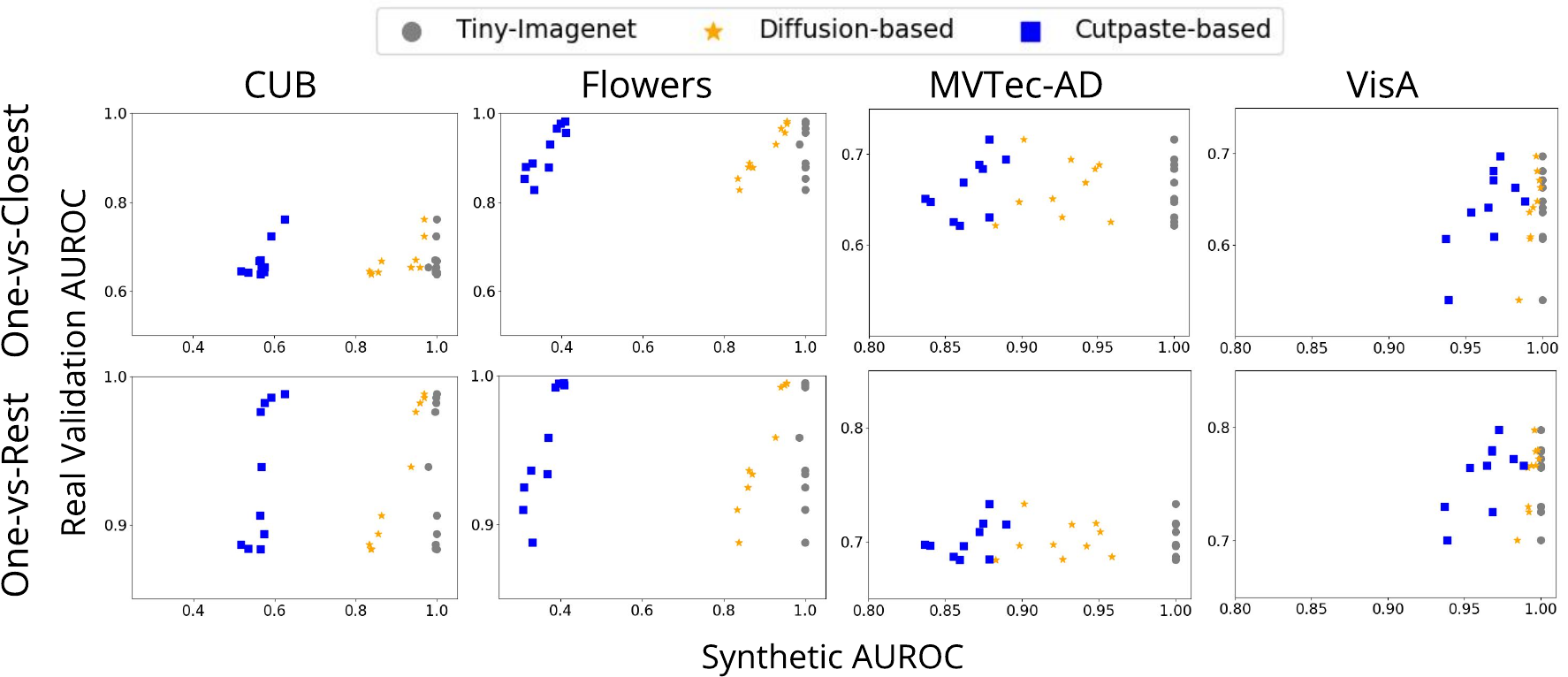}
    \caption{
        To evaluate \name~for model ranking, we compare the real and synthetic validation AUROC for all models when using our three candidate synthetic validation sets (Tiny-Imagenet, our diffusion-based anomalies, and our Cutpaste-based anomalies). 
        \name~performs best when ranking models with diffusion-based anomalies in the one-vs-rest anomaly detection setting on datasets with natural variation (i.e., CUB and Flowers). For a quantitative evaluation, we provide Kendall's Tau rank correlation values in \Cref{table:kendallrank}.}
    \label{fig:model-selection-scatter}
\end{figure*}
\begin{table*}[tb!]
    \centering
    \small
	\caption{To evaluate SWSA for model ranking for Diffusion-based and CutPaste-based anomalies, we calculate the Kendall's Tau rank correlation between the rankings with the synthetic and real validation datasets with varying sizes of synthetic validation dataset. Since we perform repeated tests on the same data, we apply Bonferroni correction to our tests and lower the significance threshold; cases with a statistically significant rank correlation \textit{(p $<$ 5.56e-3)} are \textbf{bolded}.
    We find that diffusion-based anomalies are effective for model ranking on (i) the one-vs-rest tasks for both CUB and Flowers, and (ii) one-vs-closest for the Flowers dataset.}
	\label{table:kendallrank}
	\begin{tabular}{cr | ccc}
	   \toprule
    &  & \multicolumn{3}{c}{\# of synthetic anomalies} \\
    & & All anomalies & 10 & 5 \\
    \midrule
	\multirow{4}{*}{\makecell[c]{One-vs-Closest}} 
    & Diffusion (CUB) & 0.644 \pv{9.14e-3} & 0.600 \pv{1.67e-2} & 0.600 \pv{1.67e-2} \\
    & CutPaste (CUB) & 0.377 \pv{1.55e-1} & 0.511 \pv{4.66e-2} & 0.555 \pv{2.86e-2} \\
    & Diffusion (Flowers) & \textbf{0.777 \pv{9.46e-4}} & \textbf{0.777 \pv{9.46e-4}} & \textbf{0.822 \pv{3.57e-4}} \\
    & CutPaste (Flowers) & 0.644 \pv{9.14e-3} & 0.466 \pv{7.25e-2} & 0.244 \pv{3.81e-1} \\
        \midrule
	\multirow{4}{*}{\makecell[c]{One-vs-Rest}} 
    & Diffusion (CUB) & \textbf{0.866 \pv{1.15e-4}} & \textbf{0.822 \pv{3.57e-4}} & \textbf{0.822 \pv{3.57e-4}} \\
    & CutPaste (CUB) & 0.600 \pv{1.67e-2} & \textbf{0.733 \pv{2.21e-3}} & \textbf{0.688 \pv{4.68e-3}} \\
    & Diffusion (Flowers) & \textbf{0.866 \pv{1.15e-4}} & \textbf{0.866 \pv{1.15e-4}} & \textbf{0.911 \pv{2.97e-5}} \\
    & CutPaste (Flowers) & \textbf{0.733 \pv{2.21e-3}} & 0.555 \pv{2.86e-2} & 0.333 \pv{2.16e-1} \\
    \bottomrule
\end{tabular}
\end{table*}


\paragraph{Evaluation results.}
\Cref{table:merged} shows, for different model selection strategies, how often the best model is picked (i.e., pick rate) and the resulting AUROC from the selected model. 
We find that diffusion-based synthetic anomalies and CutPaste-based synthetic anomalies pick the best model at the highest rate and produce the highest AUROC for six out of eight evaluation settings, even outperforming the strategy that always selects the largest model (ViT-H-14). 
In particular, \name~with diffusion-based anomalies selects the best model with the highest frequency for CUB and Flowers in all settings.

In addition to evaluating \name~for model selection, we also consider the performance of \name~in model ranking (i.e. beyond selecting only the best-performing true model). 
\Cref{fig:model-selection-scatter} shows the real and synthetic AUROC for all 10 models in the one-vs-closest (top) and one-vs-rest (bottom) settings. Ideally, the model ranking with synthetic data (along the x-axis) should match the model ranking with real data (along the y-axis). 

For Flowers and CUB, we observe that \name~with diffusion-based anomalies on one-vs-rest anomaly detection tasks provides the most consistent ranking. Although the number of datapoints per setting is low ($n=10$), we also provide a quantitative evaluation in \Cref{table:kendallrank} by calculating the Kendall's Tau rank correlation coefficients between the real and synthetic AUROC.\footnote{For Tiny-Imagenet, the synthetic AUROC $\approx$ 1.0 for most cases, and the rank correlation is near zero.}
To determine how many synthetic anomalies are sufficient, we vary the number of synthetic anomalies from the full set of anomalies to as few as five, keeping the anomalies with the lowest anomaly score (i.e., the most difficult anomalies).
We find that one-vs-closest anomaly detection tasks (i.e., worst-case performance) are harder to estimate with \name, but (i) model rankings with synthetic data are more highly correlated with real rankings on one-vs-rest tasks and (ii) diffusion-based anomalies are effective for Flowers in all settings.

For MVTec-AD and VisA, unlike the datasets of natural images, \name~is less effective for these datasets. The model selection results are less consistent and the correlation from model ranking is not statistically significant. These datasets contain anomalies with fine-grained industrial defects and are in general more difficult for model selection; we provide additional analysis and discussion of this phenomenon in \Cref{sec:futurework}.

\subsection{CLIP Prompt Selection with Synthetic Data}
\label{sec:prompt_selection}
The performance of CLIP-based anomaly detection models depends on the choice of prompts~\cite{jeong2023winclip,liznerski2022exposing}.
Prior works evaluate candidate CLIP prompts for zero-shot image anomaly detection on real labeled validation data, which we assume is not available in our setting. 
Thus, we next evaluate the efficacy of \name~in selecting CLIP prompts for our 329 anomaly detection tasks.

\paragraph{Zero-shot anomaly detection with CLIP.} 
We perform zero-shot image anomaly detection with CLIP using the method suggested in prior work: given an input image, we submit two text prompts and predict the class with the higher CLIP similarity to the image~\cite{radford2021learning}. We use the same backbone (``ViT-B-16-plus-240'') and data transformations as in prior work~\cite{jeong2023winclip}.
When creating prompts, we assume that the name of the inlier class is known. For CUB and Flowers, we use ``\texttt{some}'' to describe the anomaly class; for example, for the CUB dataset, if ``\texttt{red cardinal}'' is the name of the inlier class, we compare the CLIP similarities of ``\texttt{a photo of a red cardinal}'' to ``\texttt{a photo of some bird}''. 
For MVTec-AD and VisA, we use ``\texttt{with defect}'' for anomalous images; for example, if ``\texttt{transistor}'' is the name of a product, we compare ``\texttt{a photo of a transistor}'' to ``\texttt{a photo of a transistor with defect}''. 
We select amongst a pool of ten prompt templates used in prior work~\cite{jeong2023winclip} for our anomaly detection tasks. A full list of candidate prompts used for each dataset can be found in \Cref{app:prompts}.

\begin{figure*}[t!]
    \centering
    \includegraphics[width=0.95\linewidth]{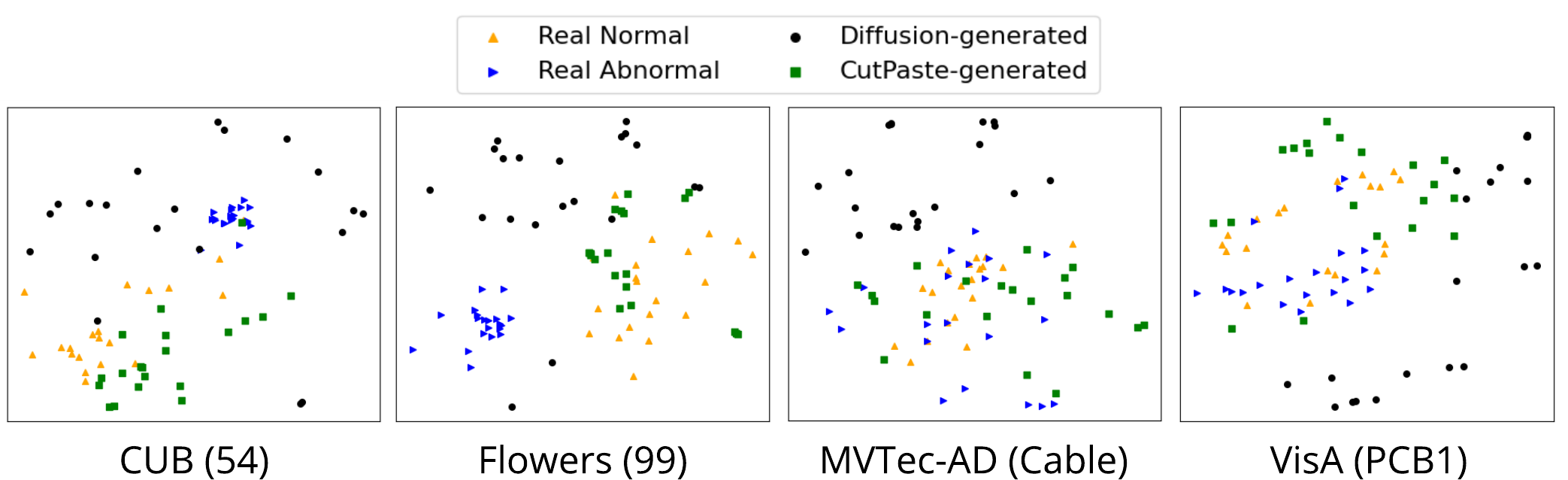}
    \caption{For each anomaly detection task, the ViT-B-16 embeddings of in-class images (orange triangle), diffusion-generated anomalies (black circle), Cutpaste-generated anomalies (green square), and real one-vs-closest anomalies (blue triangle) are shown. 
    When anomalies come from natural variations between classes (CUB and Flowers), they are better represented by diffusion-based anomalies. 
    When anomalous images come from local changes, they are better represented by Cutpaste-based anomalies. }
    \label{fig:tsne-examples}
\end{figure*}

\paragraph{Evaluation setup.}
We evaluate each candidate prompt in the one-vs-closest and one-vs-rest settings for all 329 anomaly detection tasks. We select the prompt with the best synthetic AUROC and report how often each strategy's selected prompt matches the best prompt on real validation data---we report the resulting AUROC for each strategy.
For comparison, we also include two baseline strategies: (i) the default prompt template (e.g., ``\texttt{a photo of a [class name] bird}'' vs ``\texttt{a photo of some bird}'') and (ii) a full prompt ensemble as proposed by Radford et al.~\cite{radford2021learning} and evaluated in prior work~\cite{jeong2023winclip,Zhou24}.

\paragraph{Evaluation results.} 
For all four datasets, in \Cref{table:merged}, we report the rate at which the best prompt is picked (i.e., pick rate) and the resulting AUROC of the selected prompt. We find that \name~performs best for Flowers and CUB with diffusion-based anomalies; we are able to select better prompts with our synthetic validation sets and improve the zero-shot CLIP AUROC over the popular prompt ensemble for all settings. 
We find that \name~performs best for MVTec-AD and VisA with CutPaste-based anomalies; although \name~is unable to outperform the prompt ensemble in all settings, it does outperform the prompt ensemble in the one-vs-closest setting for both MVTec-AD and VisA. This suggests that, although the prompt ensemble is more effective in general, it is not always the best strategy and \name~can be effective for prompt selection in particularly difficult settings (i.e., the worst-case anomaly detection task).
Overall, we find that our synthetic validation sets can be used to select the best prompts for CLIP-based anomaly detectors on tasks of varying domain and difficulty.

\subsection{Qualitative Results}
\label{sec:futurework}
One assumption of our work is that validation results based on synthetic anomalies can replace validation results based on labeled ground truth validation sets. 
We have shown quantitatively that synthetic anomalies provide competitive model selection strategies. 
While it would be desirable for the synthetic anomalies to reflect the underlying true distribution of anomalies, this is not an assumption we can make---anomalies, by definition, are too rare and the distribution of anomalies cannot be estimated in practice. 
Hence, in this section we qualitatively compare true anomalies with artificial anomalies. 

Figure~\ref{fig:tsne-examples} shows a t-SNE visualization of the ViT-B-16 model's embeddings for four different one-vs-closest anomaly-detection tasks---one from each dataset. 
Each plot shows the embeddings of (i) real in-class images, (ii) diffusion-generated synthetic anomalies, (iii) CutPaste-generated synthetic anomalies, and (iv) real anomalies.
Real anomalies are not available for our method in practice, but we provide their embeddings for illustration.


We find that, when anomalies come from natural variations (i.e., different species of flowers), they are further from normal data and are similar to diffusion-based anomalies. 
Conversely, when anomalies come from subtle, local changes (i.e., a defective pin on a chip), they are more tightly clustered around normal data and are similar to Cutpaste-based anomalies. 
Figure~\ref{fig:tsne-examples} shows that ground-truth anomalies for MVTec-AD and VisA are harder to distinguish from normal data and overlap with the normal data distribution, which makes detecting these anomalies with foundation models more difficult and explains why performance is worse.


Following Shoshan et al.~\cite{shoshan23synthetic}, we also perform a theoretical analysis of \name~by computing the total variation between our real and synthetic validation sets to determine if a tight bound exists for model rank preservation. Details are provided in \Cref{app:totalvar}.
We ultimately find that a tight bound does not exist, although our empirical results in \Cref{table:merged} show that \name~often selects models and prompts that match selections made with the ground-truth validation set.

\section{Conclusion}

In this work, we propose and evaluate \name: an approach for selecting image-based anomaly detection models \emph{without labeled validation data}. 
We use a general-purpose diffusion model to generate synthetic anomalies using only a small support set of in-class examples, without relying on any model training or fine-tuning.
Our empirical study shows that \name~can be used to select amongst candidate image-based anomaly detection models and to select prompts for zero-shot CLIP-based anomaly detection. 
\name~can outperform baseline selection strategies, such as using the largest model available or a prompt ensemble.

\section*{Impact Statement}
This paper presents work whose goal is to advance the field of Machine Learning. There are many potential societal consequences of our work, none of which we feel must be specifically highlighted here.

\bibliography{biblio}
\bibliographystyle{icml2024}

\newpage
\appendix
\onecolumn
\section{Prompt Templates used for CLIP-based Anomaly Detection}
\label{app:prompts}

For our experiments in \Cref{sec:prompt_selection}, we evaluated across a set of ten candidate prompt templates. 
Our evaluated prompts are general-purpose, and only the term ``bird'' or ``flower'' is added to the template for the CUB and Flowers dataset respectively. 
For MVTec-AD and VisA, we only perform mild class-name cleaning: we remove trailing numbers from class names and fully write all acronyms (e.g., ``PCB1'' becomes ``printed circuit board''). Unlike the techniques used in prior work~\cite{jeong2023winclip}, we do not perform any other class-specific modifications. 
For each dataset, the candidate prompt templates are provided below. For the results shown in \Cref{table:merged}, the first prompt listed is the default prompt, and all ten prompts are averaged for the prompt ensemble. 
\begin{tcolorbox}
\begin{footnotesize}
\begin{verbatim}
% CLIP Templates for Flowers
['a photo of a {} flower', 'a photo of some flower'],
['a cropped photo of a {} flower', 'a cropped photo of some flower'], 
['a dark photo of a {} flower', 'a dark photo of some flower'], 
['a photo of a {} flower for inspection', 'a photo of some flower for inspection'], 
['a photo of a {} flower for viewing', 'a photo of some flower for viewing'], 
['a bright photo of a {} flower', 'a bright photo of some flower'],
['a close-up photo of a {} flower', 'a close-up photo of some flower'], 
['a blurry photo of a {} flower', 'a blurry photo of some flower'], 
['a photo of a small {} flower', 'a photo of a small some flower'], 
['a photo of a large {} flower', 'a photo of a large some flower'], 
\end{verbatim}
\end{footnotesize}
\end{tcolorbox}

\begin{tcolorbox}
\begin{footnotesize}
\begin{verbatim}
% CLIP Templates for CUB
['a photo of a {} bird', 'a photo of some bird'],
['a cropped photo of a {} bird', 'a cropped photo of some bird'], 
['a dark photo of a {} bird', 'a dark photo of some bird'], 
['a photo of a {} bird for inspection', 'a photo of some bird for inspection'], 
['a photo of a {} bird for viewing', 'a photo of some bird for viewing'], 
['a bright photo of a {} bird', 'a bright photo of some bird'],
['a close-up photo of a {} bird', 'a close-up photo of some bird'], 
['a blurry photo of a {} bird', 'a blurry photo of some bird'], 
['a photo of a small {} bird', 'a photo of a small some bird'], 
['a photo of a large {} bird', 'a photo of a large some bird'], 
\end{verbatim}
\end{footnotesize}
\end{tcolorbox}

\begin{tcolorbox}
\begin{footnotesize}
\begin{verbatim}
% CLIP Templates for MVTec-AD and VisA
['a photo of a {}', 'a photo of a {} with defect'],
['a cropped photo of a {}', 'a cropped photo of a {} with defect'], 
['a dark photo of a {}', 'a dark photo of a {} with defect'], 
['a photo of a {} for inspection', 'a photo of a {} with defect for inspection'], 
['a photo of a {} for viewing', 'a photo of a {} with defect for viewing'], 
['a bright photo of a {}', 'a bright photo of a {} with defect'],
['a close-up photo of a {}', 'a close-up photo of a {} with defect'], 
['a blurry photo of a {}', 'a blurry photo of {} with defect'], 
['a photo of a small {}', 'a photo of a small {} with defect'], 
['a photo of a large {}', 'a photo of a large {} with defect'], 
\end{verbatim}
\end{footnotesize}
\end{tcolorbox}


\section{Computing Total Variation}
\label{app:totalvar}

Shoshan et al.~\cite{shoshan23synthetic} study model selection with synthetic data in the binary classification setting; 
their theoretical analysis also assumes that synthetic validation data does not come from the ground-truth data distribution. 
They show that the total variation distance between a synthetic validation set and true data provides an upper bound on the empirical risk difference between any two classifiers. 
We follow the method proposed by Sajjadi et al.~\cite{sajjadi2018assessing} to compute the total variation between synthetic and real validation datasets in this work:
given two datasets $D_1$, $D_2$, we convert them to their embedding space $F(D)$ and perform a k-means clustering on their union $F(D_1) \cup F(D_2)$. 
We assign each point to a cluster $y \in \{0 \dots k\}$ and separate the cluster assignments into $Y_1$ and $Y_2$ based on their original datasets. 
We then use the histogram of the corresponding cluster assignments to compute the total variation: $TV = \sum_i | K_{i,1} - K_{i,2} |$, where $K_{i,j}$ is the number of points assigned to cluster $i$ in dataset $j$. 
As a heuristic, we use $k = \sqrt{|D_1| + |D_2|}$ clusters when computing the total variation, and report the average of ten executions of k-means clustering. 
When computing the total variation between validation datasets, we let $D_1 = X_\text{support} \cup X_\text{out}$ and $D_2 = X_\text{support} \cup \Tilde{X_\text{out}}$, where $\Tilde{X_\text{out}}$ are the synthetic anomalies produced by our diffusion-based or CutPaste-based method.

\begin{table}[h]
	\centering
	\caption{For the classes shown in \Cref{fig:tsne-examples}, we compute the total variation between the real validation set and the synthetic validation set. Despite positive empirical results, no setting provides a tight bound, and we cannot provide strong guarantees of rank preservation on anomaly detection tasks.}
	\label{table:tv}
	\begin{tabular}{r|cccc}
	\toprule
    Real vs: & Flowers & CUB & MVTec-AD & VisA \\ 
    \midrule
    \makecell[r]{Diffusion-based} & 0.698 & 1.257 & 1.143 & 0.934 \\
    \makecell[r]{Cutpaste-based} & 0.697 & 1.042 & 0.534 & 0.804 \\
    \bottomrule
    \end{tabular}
\end{table}


Table~\ref{table:tv} shows the total variation for each dataset and type of synthetic anomaly.
Although our empirical results in \Cref{sec:ad_results} and \Cref{sec:prompt_selection} show that our synthetic anomalies can be used for model selection and prompt selection, we find that the total variation does not provide a tight bound for synthetic validation sets with Cutpaste-based nor diffusion-based anomalies (the empirical risk difference is over 0.5 in all settings).




\end{document}